\documentclass[letterpaper]{article} 
\usepackage{aaai18}  
\usepackage{times}  
\usepackage{helvet}  
\usepackage{courier}  
\usepackage{url}  
\usepackage{graphicx}  
\usepackage{amsmath}
\usepackage{booktabs}
\usepackage{multirow}
\usepackage{algorithm}
\usepackage{algorithmic}
\title{Incorporating GAN for Negative Sampling in Knowledge Representation Learning}
\author{
	Peifeng Wang \\ School of Data and Computer Science, \\ Sun Yat-sen University, China. \\wangpf3@mail2.sysu.edu.cn
	\And
	Shuangyin Li \\ iPIN Inc., Shenzhen, China. \\shuangyinli@ipin.com
	\And
	Rong Pan\thanks{Corresponding author.} \\ School of Data and Computer Science, \\ Sun Yat-sen University, China.\\ panr@sysu.edu.cn
	}

\begin{document}
\maketitle

\begin{abstract}
Knowledge representation learning aims at modeling knowledge graph by encoding entities and relations into a low dimensional space. Most of the traditional works for knowledge embedding need negative sampling to minimize a margin-based ranking loss. However, those works construct negative samples through a random mode, by which the samples are often too trivial to fit the model efficiently. In this paper, we propose a novel 
knowledge representation learning framework based on Generative Adversarial Networks (GAN). In this GAN-based framework, we take advantage of a generator to obtain high-quality negative samples. Meanwhile, the discriminator in GAN learns the embeddings of the entities and relations in knowledge graph. Thus, we can incorporate the proposed GAN-based framework into various traditional models to improve the ability of knowledge representation learning. Experimental results show that our proposed GAN-based framework outperforms baselines on triplets classification and link prediction tasks.

\end{abstract}

\section{Introduction}
Knowledge graph refers to a network whose nodes are entities in the real world and edges are relations between entities. Such a network builds up a structural system for human knowledge and is composed of a large number of facts in the form of triplet (head entity, relation, tail entity). Knowledge graph plays an important role in supporting applications such as web search, question answering, and personalized recommendation. Many knowledge graphs, such as Freebase \cite{bollacker2008freebase}, DBpedia \cite{auer2007dbpedia}, and YAGO \cite{suchanek2007yago}, have already been well-developed. Lots of previous works~\cite{bordes2013translating,wang2014knowledge,lin2015learning,he2015learning} on knowledge graphs make great efforts in knowledge representation learning to handle data sparsity and incompleteness.
Knowledge representation learning attempts to embed a knowledge graph into a continuous vector space. It provides a numerical computation framework for knowledge graph completion and captures semantic similarity between entities, which is useful for solving data sparsity.

\begin{figure}
	\includegraphics[scale=0.7]{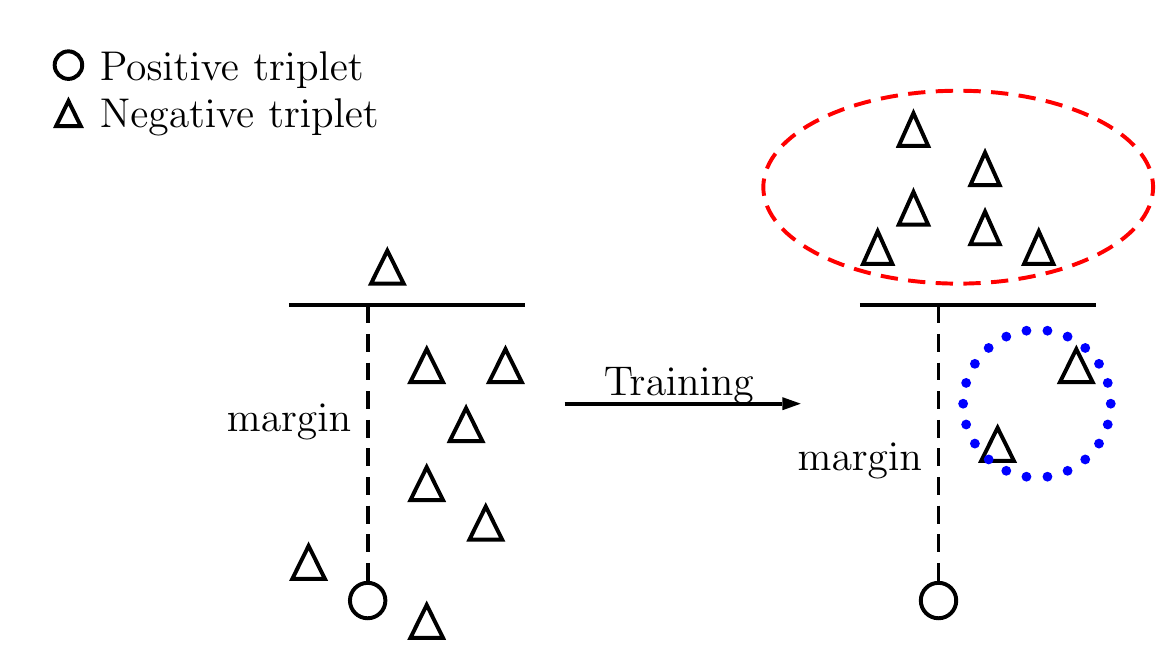}
	\caption{Illustration of zero loss problem in the random mode. After training for a while, the random mode often samples triplets in the red segmented circle, which are out of the margin and bring no loss. A certain number of negative triplets are thus ignored within the margin as shown in the blue dotted circle.}
	\label{fig:illustration}
\end{figure}

To date, many knowledge representation learning models have been proposed and have shown to achieve great successes for their efficiency. These models try different approaches to score a triplet $(h,r,t)$, which assume that there are positive triplets from the training set and man-made negative triplets. The target of these models is to minimize a margin-based ranking loss which aims to separate the scores of the positive triplets from those of the negative triplets by a fixed margin. 

Nevertheless, the negative triplets are constructed by replacing either head or tail entities of positive triplets with entities randomly sampled from the whole entities set. Such a random sampling method is effective at the beginning of the training since most of the negative triplets are still within the margin to the positive triplets. However, when the training process goes on, if we still construct negative triplets by random sampling, they are very likely to be out of the margin, which gives rise to zero loss problem as illustrated in Figure~\ref{fig:illustration}. Thus, these trivial negative triplets fail to provide guidances on model training. 

Moreover, a high-quality negative triplet can help push the model to its limit. Take the triplet (\emph{Steve Jobs, FounderOf, Apple Inc.}) as an example. We would like to replace its head entity, \emph{Steve Jobs}, to generate a negative triplet. If we just sample randomly, there is a large probability that we might sample some less informative negative entities such as \emph{London} or \emph{baseball} which are not person entity at all, resulting zero loss easily. Being told over and over again that \emph{Steve Jobs} is different from the non-person entities, the model can not learn the real concept of this person entity. Thus randomly generated negative triplets usually do little help in training the true triplets and even slows down the convergence as pointed out by \cite{schroff2015facenet,hermans2017defense}. Therefore, it is necessary to mine quality negative triplets while training.

Thus, in this paper, we propose a knowledge embedding framework based on GAN \cite{goodfellow2014generative}
to improve the negative sampling in knowledge representation learning. The discriminator in it is trained to minimize the margin-based ranking loss as in the previous models. The generator learns mining quality negative samples which can bring non-zero loss to the discriminator. The embeddings learned by the discriminator are used as the final representation of the knowledge graph. The generator could be considered as an auxiliary which pushes the discriminator to its limit in modeling the knowledge graph. Our contributions in this paper are as follows. 
\begin{enumerate}
	\item We incorporate the GAN framework in knowledge representation learning, and prove that the generator can consistently provide quality negative triplets which is crucial for the discriminator to minimize the margin-based ranking loss.
	\item The proposed GAN-based training framework can be easily extended to various knowledge representation learning models and enhance their ability in knowledge embedding tasks.
\end{enumerate}

We evaluate the proposed GAN-based training framework on two knowledge graph related tasks, triplets classification and link prediction.  The experimental results show that the performances of our GAN-based models can achieve remarkable improvements over the baselines, which justifies the effectiveness of our model in mining quality negative samples and embedding knowledge graph.

\section{Related Work}\label{rw}
For symbol clarification, we define an entity set \textbf{E} and a relation set \textbf{R}. We denote a triplet as (h, r, t) where h, t represent head and tail entity respectively, and r indicates the relation between h and t. The corresponding embeddings we would like to learn are \textbf{h}, \textbf{t} and \textbf{r} for entities h, t and relation r. 

\textbf{TransE}~\cite{bordes2013translating} starts the line of translation based knowledge embedding models and the basic idea of these models is that if a triplet (h, r, t) holds, then it represents a translation from a head entity to a tail entity via a relation vector, namely $\textbf{h} + \textbf{r} = \textbf{t}$. The score function used by TransE for a triplet (h, r, t) is defined as
\begin{equation}
f_r(h, t) = \vert \mathbf{h} + \mathbf{r} - \mathbf{t}\vert_{L1/L2},
\end{equation}
where L1 and L2 represent L1 and L2 norm. The model should achieve lower score if a triplet holds.

Since TransE is not suitable for 1-to-N, N-to-1 or N-to-N relations, \textbf{TransH} \cite{wang2014knowledge} proposes to project the entity to the relation specific hyperplane $\mathbf{w_r}$ and thus an entity could have different representations depending on what relation evolved. In detail, TransH transforms the entity embeddings as
\begin{equation}
	\mathbf{h}_\perp=\mathbf{h}-\mathbf{w_r}^\top\mathbf{h}\mathbf{w_r},
	\quad\mathbf{t}_\perp=\mathbf{t}-\mathbf{w_r}^\top\mathbf{t}\mathbf{w_r},
\end{equation}
and the score function is defined as
\begin{equation}
	f_r(h, t) = \vert \mathbf{h}_\perp + \mathbf{r} - \mathbf{t}_\perp\vert_{L1/L2}.
\end{equation}

Both TransE and TransH assume that entities and relations are in the same vector space, thus \textbf{TransR} \cite{lin2015learning} proposes to use a relation specific matrix $\mathbf{M_r}$ to project the entity embedding into the relation vector space and the score function becomes
\begin{equation}
	f_r(h, t) = \vert\mathbf{M_r}\mathbf{h} + \mathbf{r}-\mathbf{M_r}\mathbf{t}\vert_{L1/L2}.
\end{equation} 

\textbf{TransD} \cite{ji2015knowledge} argues that the project matrix should not only be relation specific, but also be determined by entities themselves. Therefore the project matrices are defined as
\begin{equation}
	\mathbf{M_{rh}}=\mathbf{r_p}\mathbf{h_p^\top}+\mathbf{I},
	\quad\mathbf{M_{rt}}=\mathbf{r_p}\mathbf{t_p^\top}+\mathbf{I},
\end{equation} 
where subscript $p$ marks the projection vectors.

Other translation embedding models project entities in different ways like~\cite{xiao2015one,ji2016knowledge}  or make use of supplement information including text descriptions \cite{zhong2015aligning,xie2016representation,ijcai2017-0183} and images \cite{xie2016image} of entities.

\textbf{Unstructured Model} \cite{bordes2012joint,bordes2014semantic} ignores the relation between entities and the score function is defined as
\begin{equation}
h_r(h,t)=\vert\mathbf{h}-\mathbf{t}\vert.
\end{equation} 

\textbf{Structured Embedding} \cite{bordes2011learning} uses two projection matrix $\mathbf{M_{r,h}}$ and $\mathbf{M_{r,t}}$ to represent a relation r. The score function is defined as
\begin{equation}
f_r(h,t)=\vert\mathbf{M_{r,h}}\mathbf{h}-\mathbf{M_{r,t}}\mathbf{t}\vert.
\end{equation} 

\textbf{Single Layer Model} \cite{socher2013reasoning} improves Structured Embedding Model by defining additional relation vector $\mathbf{u_r}$ for each relation and nonlinear transformation. Its score function for a triplet is
\begin{equation}
f_r(h,t)=-\mathbf{u_r}^\top g(\mathbf{M_{r,h}}\mathbf{h}+\mathbf{M_{r,t}}\mathbf{t}+\mathbf{b_r}),
\end{equation}  
where g() is tanh function. 

\textbf{Neural Tensor Network} \cite{socher2013reasoning} extends Single Layer Model by considering the second-order correlations into non-linear transformation. Its score function is defined as 
\begin{equation}
f_r(h,t)=-\mathbf{u_r}^\top g(\mathbf{h}^\top\mathbf{M_r}\mathbf{t}+\mathbf{M_{r,h}}\mathbf{h}+\mathbf{M_{r,t}}\mathbf{t}+\mathbf{b_r}),
\end{equation}
where $\mathbf{M_r}$ is a three-way tensor.

All the above knowledge embedding models construct negative triplets by random sampling and are trained to separate the scores between positive and negative triplets. In this paper, we propose to incorporate GAN training framework for better negative sampling. GAN has been applied to many others scenarios, such as selecting difficult documents in information retrieval \cite{wang2017irgan} and generating fake sentences for commonsense machine comprehension \cite{ijcai2017-0576}. We are the first to employ GAN on knowledge representation learning for high-quality negative triplets sampling.

\section{GAN on Knowledge Embedding}
Given a training set S of triplets $(h,r,t)$, we try to learn the embeddings of entities and relations. Consider there's a positive triplet $(h,r,t)$ that holds in knowledge graph and a negative triplet $(h^{'},r,t^{'})$ that we need to construct. To ensure that the model gives a low score if a triplet (h, r, t) holds and a high score otherwise, the margin-based ranking loss is considered as
\begin{equation}\label{eq:hinge_loss}
	l(h,r,t) = [f_r(h,t) - f_r(h^{'},t^{'}) + \gamma]_+,
\end{equation}
where $[x]_+=max\{0,x\}$ denotes the positive part of x, and $\gamma$ is a fixed margin set as hyper parameter. The score function $f_r(h,t)$ has been described in Section~\ref{rw}. The objective function of the model is defined as
\begin{equation}\label{eq:object_function}
\min\sum\limits_{(h,r,t)\in\Delta}
			\sum\limits_{(h^{'}, r,t^{'})\in\Delta^{'}}l(h,r,t),
\end{equation}
where the $\Delta$ denotes the positive triplets set from the knowledge graph and $\Delta^{'}$ indicates the negative triplets set constructed by replacing the head or tail entity of the triplets in $\Delta$ by other entity in \textbf{E}, namely
\begin{equation}
\Delta^{'}=\{(h^{'},r,t)|h^{'}\in\textbf{E}\}\cup\{(h,r,t^{'})|t^{'}\in\textbf{E}\}.
\end{equation}
In practice, the model enforces constrains as $\Vert \mathbf{h}\Vert_2\leq 1$, $\Vert \mathbf{r}\Vert_2\leq 1$ and $\Vert \mathbf{t}\Vert_2\leq 1$, where $\Vert \cdot\Vert_2$ refers to L2 norm.

During training, positive triplets are traversed randomly. When a positive triplet is visited, previous methods generate a negative triplet by replacing the head entity or tail entity (not both at the same time) with an entity sampled randomly from the whole entity set $\mathbf{E}$. There are two strategies to decide whether to replace head or tail entity, (1) ``\textbf{unif}'' method which replaces head or tail with equal probability; (2) ``\textbf{bern}'' method from \cite{wang2014knowledge} which replaces head or tail according to Bernoulli distribution.

\textbf{Zero loss problem.} The negative triplets generated by random sampling are effective at the beginning of training. However, after training for a while, most of these randomly generated negative triplets achieve scores that are out of the margin to the positive triplets and would lead to zero loss in Eq.~(\ref{eq:hinge_loss}) trivially. These negative triplets make no contribution to improving the embeddings. Thus such a sampling method would cause very slow convergence and even fail to get the best result. 

To generate quality negative triplets and accelerate the training process, we propose to incorporate the GAN framework in our model, under which the discriminator is trained to minimize the margin-based ranking loss in Eq.~(\ref{eq:hinge_loss}) as the previous models, while the generator learns to continually generate quality negative triplets that would not lead to zero loss. We describe the discriminator and generator respectively in Section~\ref{d} and Section~\ref{g}.

\subsection{Discriminator for Knowledge Embedding}\label{d}
The discriminator in our model is designed as the previous models. For different models, the score function is computed in various ways as introduced in Section~\ref{rw}. Then the embeddings of entities and relations are learned by training the discriminator to minimize the margin-based ranking loss between positive triplets and negative triplets as defined in Eq.~(\ref{eq:hinge_loss}). Different to the previous models whose negative triplets are constructed by randomly sampling from the whole entity set \textbf{E}, the discriminator uses the negative triplets generated by the generator which will be described in detail as in Section~\ref{g}.

\subsection{Generator for Negative Sampling}\label{g}
The goal of the generator is to provide the discriminator with quality negative triplets which could bring non-zero loss. Note that, the generator in our model has its own embeddings (different from discriminator's) for entities and relations. Besides, there are two separate embeddings for each relation $r$, one for normal $r$ as usual and another for the reverse of the relation, namely $r^{-1}$. For each positive triplet $(h, r, t)$, the input of the generator is an entity-relation pair defined as
\begin{equation}
\begin{cases}
(t,r), &\quad z = 1\\
(h,r^{-1}), &\quad z = 0,
\end{cases}
\end{equation}
where $z\in\{1, 0\}$ is a binary flag indicating whether to replace head entity or tail entity, which is decided by either ``\textbf{unif}'' or ``\textbf{bern}'' strategy. 

\begin{figure}
	\centering
	\includegraphics[scale=0.58]{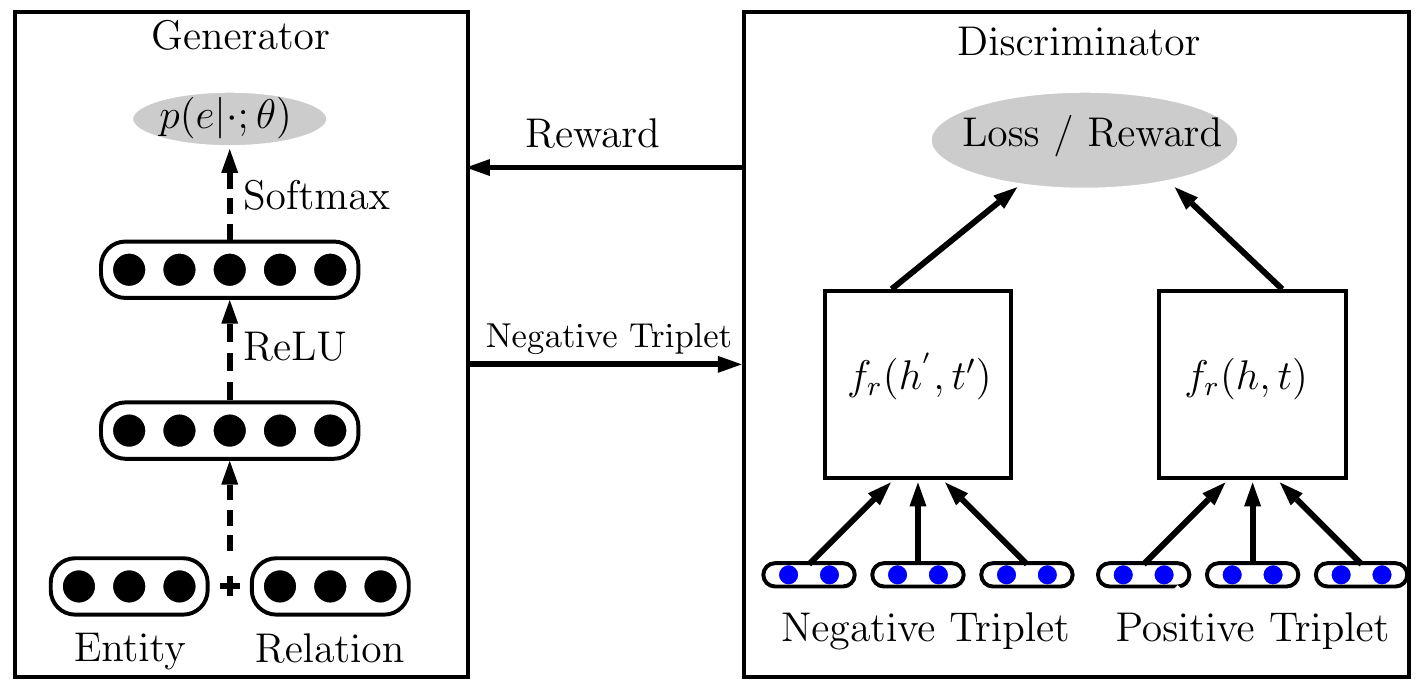}
	\caption{GAN-based training framework. (a) Generator learns to provide quality negative triplets according to the reward from discriminator. (b) Discriminator learns to embed the knowledge graph using the negative triplets from generator. }
	\label{fig:policy_network}
\end{figure}

The embeddings of the entity-relation pair are concatenated and fed to a two layers fully-connected neural network as shown in the ``Generator'' part of Figure~\ref{fig:policy_network}. Non-linear function ReLU is added after the first layer and Softmax function is added to the second layer. The network and the generator's embedding matrices for entities and relations are used to parametrize the probability distribution over the whole entity set \textbf{E}. The probability distribution of the entity set \textbf{E} is defined as
\begin{equation}
\begin{aligned}
p(e|(h,r,t),z;\theta) &= z\cdot p(e|t,r;\theta) \\ &+ (1-z)\cdot p(e|h, r^{-1};\theta),
\end{aligned}
\end{equation} 
from which we sample an entity to be the replacement. Since the output of the generator is a discrete index of the entity, we use policy gradient based reinforcement learning \cite{williams1992simple,sutton2000policy} and thus the whole network could be viewed as the policy network in the field of reinforcement learning. For a newly generated negative triplet $(h^{'}, r, t)$ or $(h, r, t^{'})$ together with the positive triplet $(h, r, t)$, the reward function calculating by the discriminator is defined as
\begin{equation}\label{eq:reward_function}
	R = tanh(f_r(h, t) - f_r(h^{'}, t^{'}) + \gamma), 
\end{equation}
where the inner part of the activation function $tanh(\cdot)$ is similar to the ranking loss used in Eq.~(\ref{eq:hinge_loss}) except that the negative part is kept. Such a reward would be positive when the generated negative triplet together with the corresponding positive triplet achieve a non-zero loss in Eq.~(\ref{eq:hinge_loss}) and negative otherwise. 

The generator is trained to maximize the expected reward 
\begin{equation}\label{eq:expected_reward}
	J(\theta)=E_{e\sim p(e|\cdot;\theta)}[R],
\end{equation}
from which we could see that in order to achieve higher reward, the policy used by the generator would punish the trivial negative entities by lowering down their corresponding probability and encourage the network to distribute more probability to the entities that can bring non-zero loss.

The generator is updated using the policy gradient as
\begin{equation}\label{eq:policy_gradient}
	\nabla_\theta J(\theta)=\nabla_\theta \log p(e|(h,r,t),z;\theta)\cdot R.
\end{equation}

\subsection{Model Training}
The training process of our proposed model is summarized in Algorithm~\ref{Algo:GAN}. To be specific, at each training epoch we iterate over the training set in mini-batch to train the generator while the parameters of the discriminator are fixed. Then we iterate over the training set again to train the discriminator while the parameters of the generator are fixed. This training process could also be regarded as a way to make the generator search for the quality negative triplets and filter out the less informative ones before we use it to challenge the discriminator. Both the discriminator and generator are trained by Adam stochastic optimization~\cite{kingma2014adam}. In addition, we apply $L_2$ regularization on generator's parameters, namely $\theta$. When the model converges, we take the embeddings learned by the discriminator as our final representation for entities and relations.

\begin{algorithm}[h]
	\caption{GAN for knowledge representation learning}
	\label{Algo:GAN}
	\renewcommand{\algorithmicrequire}{\textbf{Input:}}
	\renewcommand{\algorithmicensure}{\textbf{Output:}}
	\begin{algorithmic}[1]
		\REQUIRE Generator G, Discriminator D; \\
		\hspace*{5mm} Train set $\mathcal{S}=\{(h,r,t)\}$
		\ENSURE Knowledge representations learned by D
		\LOOP
		\FOR{g steps}
		\STATE Sample a batch of positive triplets from $\mathcal{S}$
		\STATE Use G to generate negative triplets
		\STATE Use D to calculate REWARD in Eq.~(\ref{eq:reward_function})
		\STATE Update G's parameters via policy gradient in Eq.~(\ref{eq:policy_gradient})
		\ENDFOR
		\FOR{d steps}
		\STATE Sample a batch of positive triplets from $\mathcal{S}$
		\STATE Use G to generate negative triplets
		\STATE Use D to calculate LOSS in Eq.~(\ref{eq:hinge_loss})
		\STATE Update D's parameters w.r.t Eq.~(\ref{eq:object_function})
		\ENDFOR
		\ENDLOOP
	\end{algorithmic}
\end{algorithm}

\section{Experiments and Analysis}
Experiments are conducted on two knowledge graphs, Freebase \cite{bollacker2008freebase} and Wordnet \cite{miller1995wordnet} for two reasoning tasks, link prediction and triplets classification. Freebase contains common facts of the world like (\emph{louis\_duke\_of\_brittany, place\_of\_birth, france}), and we use two subsets of Freebase. They are FB15K used in \cite{bordes2013translating} and FB13 used in \cite{socher2013reasoning}. Wordnet provides lexical relations between words like (\emph{\_discuss\_2, \_type\_of, \_talk\_about\_2}) and we also use two subsets of Wordnet. They are WN11 used in \cite{socher2013reasoning} and WN18 used in \cite{bordes2013translating}. The statistics are listed in Table~\ref{tab:dataset_statistics}.

\begin{table}[h]
	\centering
	\caption{Data sets used in the experiments.}\label{tab:dataset_statistics}
	\begin{tabular}{c|ccccc}
		\toprule
		Dataset & \#Rel & \#Ent & \#Train & \#Valid & \#Test \\
		\midrule
		FB15K & 1,345 & 14,951 & 483,142 & 50,000 & 59,071\\
		FB13 & 13 & 75,043 &  316,232 & 5,908 & 23,733\\
		WN11 & 11 & 38,696 & 112,581 & 2,609 & 10,544\\
		WN18 & 18 & 40,943 & 141,442 & 5,000 & 5,000\\
		\bottomrule
	\end{tabular}
\end{table}

We compare the proposed GAN-based knowledge embedding framework with the models introduced in Section~\ref{rw}. Two training settings are conducted as follows.
\begin{itemize}
\item \textbf{GAN-scratch} The parameters of the discriminator and generator are initialized randomly. Thus the whole model is trained from scratch.
\item \textbf{GAN-pretrain} We firstly train the original models with random negative sampling. Then the embeddings learned by the models are used to initialize the parameters of the discriminator. The parameters of the generator are still initialized randomly. Thus the whole model is trained to fine-tune the original models.
\end{itemize}

\subsection{Link Prediction}\label{lp}
Link prediction aims to predict the missing entity in a triplet, namely to predict h given (r, t) or t given (h, r). It reflects the model's ability to do knowledge reasoning based on the embeddings it learns. Instead of predicting the golden entity, we list the ranked candidate entities for each incomplete triplet. The experiments are conducted on the data sets WN18 and FB15K. The models we would like to enhance with GAN are listed in Table~\ref{tab:result1}.

\begin{table*}[t]
	\small
	\centering
	\caption{Evaluation results on link predictions.}\label{tab:result1}
	\begin{tabular}{c|cc|cc|cc}
		\toprule
		\multirow{2}{*}{FB15K} & \multicolumn{2}{c|}{Original} & \multicolumn{2}{c|}{GAN-scratch} & \multicolumn{2}{c}{GAN-pretrain}\\
		{} & Mean Rank & Hits@10 (\%) & Mean Rank & Hits@10 (\%) & Mean Rank & Hits@10 (\%) \\
		\midrule
		Unstructured~\cite{bordes2012joint} & 979 & 6.3 & 332 & 24.5 & \textbf{305} & \textbf{28.7} \\
		SE~\cite{bordes2011learning} & \textbf{162} & 39.8 & 207 & 53.3 & 220 & \textbf{53.8} \\
		SME(Bilinear)~\cite{bordes2012joint} & 158 & 41.3 & \textbf{139} & \textbf{45.8} & 170 & 45.7 \\
		TransE~\cite{bordes2013translating} & \textbf{61} & 69.6 & 90 & 73.1 & 81 & \textbf{74.0} \\
		TransH~\cite{wang2014knowledge} & 87 & 64.4 & 90 & 73.3 & \textbf{81} & \textbf{77.0} \\
		TransR~\cite{lin2015learning} & \textbf{77} & 68.7 & 138 & 58.3 & 86 & \textbf{76.0} \\
		TransD~\cite{ji2015knowledge} & 91 & 77.3 & 89 & 74.0 & \textbf{79} & \textbf{77.6}\\
		TransE + A-LSTM~\cite{ijcai2017-0183} & 77 & 75.5 & 79 & 75.8 & \textbf{74} & \textbf{76.3} \\
		\bottomrule
		\toprule
		\multirow{2}{*}{WN18} & \multicolumn{2}{c|}{Original} & \multicolumn{2}{c|}{GAN-scratch} & \multicolumn{2}{c}{GAN-pretrain}\\
		{} & Mean Rank & Hits@10 (\%) & Mean Rank & Hits@10 (\%) & Mean Rank & Hits@10 (\%) \\
		\midrule
		Unstructured~\cite{bordes2012joint} & \textbf{304} & 38.2 & 477 & 80.5 & 431 & \textbf{82.9} \\
		SE~\cite{bordes2011learning} & 985 & 80.5 & 773 & 86.1 & \textbf{762} & \textbf{87.4} \\
		SME(Bilinear) & 509 & 61.3 & \textbf{358} & 74.6 & 417 & \textbf{78.9} \\
		TransE~\cite{bordes2013translating} & 251 & 89.2 & 244 & \textbf{92.7} & \textbf{240} & 91.3 \\
		TransH~\cite{wang2014knowledge} & 303 & 86.7 & 276 & 86.9 & \textbf{258} & \textbf{94.0} \\
		TransR~\cite{lin2015learning} & 225 & 92.0 & \textbf{213} & 87.3 & 291 & \textbf{93.7} \\
		TransD~\cite{ji2015knowledge} & 229 & 92.5  & \textbf{221} & 93.0 & 248 & \textbf{93.3} \\
		TransE + A-LSTM~\cite{ijcai2017-0183} & 123 & 90.9 & \textbf{114} & 92.3 & 120 & \textbf{92.7} \\
		\bottomrule
	\end{tabular}
\end{table*}

For each test triplet (h, r, t), we replace the head (or tail) entity by all the entities in the entities set E. Then we compute the score function $f_r(h,t)$ for the test triplet and its corresponding corrupted triplets. Next we rank the scores in a descending order. We report two measures as the previous studies on knowledge representation learning, the average rank of the correct entities (Mean Rank) and the proportion of the correct entities ranked in top 10 (Hits@10). Lower Mean Rank or higher Hits@10 reflects better results. Since a corrupted triplet might also exist in the knowledge graph, ranking it ahead of the original triplet is also acceptable. Thus we follow previous studies and filter out all the corrupted triplets that exist in train, validation or test set in order to avoid underestimating the performance of our proposed model, which is called as ``Filter'' setting.

\textbf{Implementation.}
We use Adam SGD for optimization. We select the margin $\gamma$ among \{0.5, 1.0, 2.0, 4.0\}, the dimension d of entities and relations among \{20, 50, 100, 200\}, learning rate $\lambda$ for SGD among \{0.01, 0.001, 0.0001, 0.00005\} and the batch size B among \{512, 1024, 2048, 4096\}. We search the best configuration according to the performance of the model in validation set. For ``GAN-scratch'', the optimal configurations are $\gamma = 2.0$, $d=100$, $\lambda=0.001$, and $B = 1024$ on WN18; $\gamma = 1.0$, d = 100, $\lambda = 0.0001$, and $B = 4096$ on FB15K. For ``GAN-pretrain'', the optimal configurations are $\gamma = 2.0$, d = 100, $\lambda=0.00005$, and $B = 1024$ on WN18; $\gamma = 1.0$, $d=100$, $\lambda = 0.0001$, and $B = 2046$ on FB15K. For both training settings, wo adopt the L1 as dissimilarity and use the ``unif'' strategy to decide whether to replace head or tail during training.

\textbf{Result.} 
Evaluation results are reported in Table~\ref{tab:result1}. Since the datasets and the evaluation protocol are the same, we choose to reprint the best ``Filter'' experimental results of the original models directly from the literature, which are referred as ``Original'' in Table~\ref{tab:result1}, if our own implementations do not yield better results than the ones reported in the literature. Two training settings of our proposed model are referred as ``GAN-scratch'' and ``GAN-pretrain'' respectively. We can conclude from Table~\ref{tab:result1} as follows.
\begin{enumerate}
\item For most of the knowledge embedding models, our proposed GAN training framework yields a better performance on FB15K and WN11 respectively. This illustrates the effectiveness of our GAN framework for knowledge representation learning, due to the fact that the generator could generate more quality negative triplets than random sampling does. These non-trivial negative triplets push the discriminator to better rank the correct positive triplets. We also justify this point by visualizing the negative triplets generated by our generator in Section~\ref{visual}.
\item Most of the models trained under the ``GAN-pretrain'' setting perform better than their counterparts trained under the ``GAN-scratch'' setting. This phenomenon is also observed in triplets classification task, which would be discussed further in the Section~\ref{discuss}. 
\item The proposed GAN-based framework does not work significantly well on TransR and TransD model. Actually, we find it hard to train TransR with random negative sampling as well. It is easy for TransR to over-fit train dataset since TransR has much more parameters and the consistent non-zero losses brought by our GAN framework worsen this problem by pushing the model too hard. TransD also has the over-fitting problem since TransD distributes a projection vector for each entity and each relation which is not appropriate in terms of generalization. We initial TransR and TransD with the entity and relation embeddings learned from TransE as suggested in \cite{lin2015learning,ji2015knowledge} but still fail to get a noticeable improvement.
\end{enumerate}

\begin{table*}[h]
	
	\centering
	\caption{Evaluation results about classification accuracy on triplets classification(\%).}\label{tab:result2}
	\begin{tabular}{c|c|cc|c|cc}
		\toprule
		{} & \multicolumn{3}{c|}{FB13} & \multicolumn{3}{c}{WN11} \\
		{Model} & Original & GAN-scratch & GAN-pretrain & Original & GAN-scratch & GAN-prertain \\
		\midrule
		SE~\cite{bordes2011learning} & 75.2 & 84.2 & \textbf{84.7} & 53.0 & 60.1 & \textbf{63.2} \\
		SME(bilinear)~\cite{bordes2012joint} & 63.7 & 77.0 & \textbf{79.6} & 70.0 & 72.1 & \textbf{74.3} \\
		SLM~\cite{socher2013reasoning} & 85.3 & \textbf{85.6} & 83.8 & 69.9 & 72.7 & \textbf{75.3} \\
		TransE~\cite{bordes2013translating} & 81.5 & 83.1 & \textbf{85.2} & 75.9 & 85.1 & \textbf{85.4} \\
		TransH~\cite{wang2014knowledge} & 83.3 & 85.0 & \textbf{86.3} & 78.8 & 85.5 & \textbf{85.9} \\
		TransR~\cite{lin2015learning} & 82.5 & 85.4 & \textbf{86.6} & 85.9 & 85.2 & \textbf{86.3} \\
		TransD~\cite{ji2015knowledge} & 89.1 & 85.3 & \textbf{89.7} & \textbf{86.4} & 82.6 & 84.0 \\
		\bottomrule
	\end{tabular}
\end{table*}

\subsection{Triplets Classification}
Triplets classification is to classify whether a triplet (h, r, t) holds or not, which is a binary classification problem. It is also used in previous studies to evaluate the embeddings learned by the models. 

The experiments are conducted on the WN11 and FB13 datasets released by \cite{socher2013reasoning} which already have a negative triplet for each positive triplet in the validation set and test set. The negative triplets are constructed by randomly corrupting the entities of the positive triplets but with the constraint that an entity is chosen as the replacement only if it appears at the same position (head or tail) in the dataset. Thus it is not a trivial task for knowledge representation learning.

For triplets classification, we preset a threshold $\delta_r$ for each relation r. If the score of (h, r, t) is below $\delta_r$, then the triplet is classified as positive, otherwise negative. We determine the optimal $\delta_r$ by maximizing the classification accuracy on the validation set.

\textbf{Implementation.} We use Adam SGD for optimization. We select the margin $\gamma$ among \{0.5, 1.0, 2.0, 4.0\}, the dimension d of entities and relations among \{20, 50, 100, 200\}, learning rate $\lambda$ for SGD among \{0.01, 0.001, 0.0001, 0.00005\} and the batch size B among \{128, 512, 1024, 2048, 4096\}. We search the best configuration according to the accuracy on validation set. For ``GAN-scratch'', the optimal configurations are $\gamma = 4.0$, $d = 50$, $\lambda = 0.001$, and $B = 1024$ on WN11; $\gamma = 1.0$, $d = 100$, $\lambda = 0.0001$, and $B = 4096$ on FB13. For ``GAN-pretrain'', the optimal configurations are $\gamma = 4.0$, $d = 50$, $\lambda = 0.0001$, and $B = 512$ on WN11; $\gamma = 1.0$, $d = 100$, $\lambda = 0.00005$, and $B = 1024$ on FB13. For both datasets, wo adopt the L1 as dissimilarity and use the ``bern'' strategy for deciding whether to replace head or tail when training.

\textbf{Result.} Evaluation results are reported in Table~\ref{tab:result2}. We still directly reprint the best ``Filter'' experimental results of the original models from the literature, which are referred as ``Original'' in the Table~\ref{tab:result2}. The results show that
\begin{enumerate}
\item Models including SE, SME(bilinear), TransE and TransH are further improved under the ``GAN-scratch'' and ``GAN-pretrain'' settings, which demonstrates the effectiveness of our method on simple models. Simple models generalize well to test set, and the consistent non-zero losses brought by the generator in our GAN framework make sure that the models fit the training set sufficiently.
\item For sophisticated models like SLM, TransR and TransD, their improvements brought by our method are different. SLM is improved under the ``GAN-scratch'' setting on FB13 and WN11 datasets while ``GAN-pretrain'' does not work well on FB13 dataset. TransR is improved under the ``GAN-pretrain'' setting on FB13 and WN11 datasets while the ``GAN-scratch'' only yields a better result than the ``Original'' mode on the FB13 dataset. TransD is only improved under the ``GAN-pretrain'' setting on the FB13 dataset. This is due to the complexity and over-fitting problem discussed in previous Section~\ref{lp}, but still demonstrates the ability of our generator to bring non-zero losses continually.
\item Most of the models trained under the ``GAN-pretrain'' setting outperform those trained under the ``GAN-scratch'' setting on both datasets, which is consistent with the result of link prediction task in Section~\ref{lp}.
\end{enumerate}

\subsection{Discussion on GAN-scratch vs. GAN-pretrain}\label{discuss}
Both link prediction and triplets classification tasks show that the ``GAN-pretrain'' setting achieves better performance than the ``GAN-scratch'' setting under most circumstances. One reason is that the pre-trained models have already converged to some less-optimal state but fail to make further progress because of the zero loss problem. Therefore GAN is able to fine-tune them from a better starting point. Another is that the search space of entities that could bring non-zero loss for the ``GAN-pretrain'' setting is much smaller than that for the ``GAN-scratch'' setting. After pre-training, most of the negative triplets have been within the margin. Thus the policy network used by the generator in the ``GAN-pretrain'' setting might face a more stable environment while the one used in ``GAN-scratch'' face an environment that changes dynamically. The ``GAN-scratch'' setting has to gradually narrow down the search space from the whole entity set to a few of entities due to the changing reward given by the discriminator in Eq.~(\ref{eq:reward_function}). ``GAN-pretrain'' overcome this issue by enabling the policy network in generator directly search the entities within the margin since the reward given by the discriminator is already stable.

\begin{table}[h]
	\centering
	\caption{Examples of Negative Triplets from Generator on FB13. In each cell, the first row displays the original triplets and the second row displays the head and tail entity generated by our model.}\label{tab:examples}
	\begin{tabular}{lll}
		\toprule
		(\emph{william\_preyer}, & \emph{gender}, & \emph{male})\\ 
		\emph{ho\_yuen\_hoe} & {} &\emph{female}\\
		\midrule
		(\emph{ole\_rolvaag}, & \emph{religion}, &	\emph{lutheranism})\\ 
		\emph{vasil\_levski} & {} & \emph{catholicism}\\
		\midrule
		(\emph{parameshvara}, & \emph{nationality}, & \emph{india})\\
		\emph{johann\_fussli} & {} & \emph{united\_states}\\
		\midrule
		(\emph{peter\_scott}, & \emph{institution}, & \emph{royal\_academy})\\
		\emph{lester\_b\_pearson} & {} & \emph{eton\_college}\\
		\midrule
		(\emph{michel\_chartrand}, & \emph{profession}, & \emph{politician})\\
		\emph{louis\_jordan} & {} & \emph{writer}\\
		\midrule
		(\emph{mark\_fidrych}, & \emph{cause\_of\_death}, & \emph{accident})\\
		\emph{noah\_rosenzweig} & {} & \emph{cancer}\\
		\bottomrule
	\end{tabular}
\end{table}

\subsection{Visualization of Negative Triplets}\label{visual}
We visualize some typical negative triplets generated by our generator in Table~\ref{tab:examples}. We conduct the experiment on FB13 dataset since its triplets are more interpretable. The positive triplets from the training set are listed at the first row of each cell in Table~\ref{tab:examples} and the second row lists out the head and tail negative entities generated by our model. We use generator trained under the ``GAN-pretrain'' setting. From Table~\ref{tab:examples} we can see that the negative head entities are basically person entities as we expect. The negative tail entities are also quality and informative since each of them is of the same entity type as their positive counterpart respectively. As an example, for the triplet (\emph{michel\_chartrand, profession, politician}), the negative entities generated by our generator is \emph{louis\_jordan} (a person name) for the head and \emph{writer} (also a type of profession) for the tail. This justifies the ability of our generator to generate negative triplets that are more challenging for the discriminator than random sampling, which pushes our discriminator to its limit in embedding entities and relations. Thus, our model can learn the real concept of each entity better since the discriminator is trained to distinguish the positive entities from the high-quality negative entities.

\section{Conclusion and Future Work}
We proposed to incorporate the GAN training framework for knowledge representation learning. The generator could provide much more quality negative triplets than random sampling while the discriminator thus could be helped to learn better embeddings for entities and relations. We show by extensive experiments that the GAN-based framework could efficiently improve several existing knowledge embedding models and the experimental results show that the idea can generalize well to the state-of-the-art knowledge embedding models.

\textbf{Future work.} We will conduct additional analysis on our GAN-based framework for knowledge representation learning and its generalization to other problems that involve negative sampling. Also, we will experiment with a more complicated architecture of the generator.

\section{Acknowledgements}
This work was supported by the National Key R\&D Program of China under Grant 2016YFB0201900, and the Fundamental Research Funds for the Central Universities under Grant 17LGJC23. This work was also supported by iPIN Inc. Shenzhen, China.

\bibliographystyle{aaai}

\end{document}